\documentclass{article} 
\usepackage{iclr2027_conference,times}

\usepackage{amsmath,amsfonts,bm}

\def\eqref#1{equation~\ref{#1}}

\def\1{\bm{1}}

\DeclareMathAlphabet{\mathsfit}{\encodingdefault}{\sfdefault}{m}{sl}
\SetMathAlphabet{\mathsfit}{bold}{\encodingdefault}{\sfdefault}{bx}{n}

\usepackage{hyperref}
\usepackage{url}
\usepackage{csquotes}
\usepackage{booktabs}
\usepackage[utf8]{inputenc}
\usepackage[T1]{fontenc}
\usepackage[noend]{algorithm2e} 
\usepackage{amsmath}
\usepackage{amsfonts}
\usepackage{bbm}
\usepackage{wrapfig}
\usepackage[dvipsnames]{xcolor}
\usepackage{comment}
\usepackage[labelfont=bf]{caption}
\usepackage{tcolorbox}
\usepackage{enumitem}
\usepackage{subcaption}
\usepackage{tikz} 
\usetikzlibrary{tikzmark}

\newcommand{\franz}[1]{\textcolor{blue}{[Franz: #1]}}
\newcommand{\teo}[1]{\textcolor{magenta}{[Teo: #1]}}

\title{Verifying the Linear Representation Hypothesis: 
How Interpretable Are Vision SAEs?}

\author{%
    Teodor Chiaburu$^{1, 2}$\thanks{Correspondence: \texttt{teodor-constantin.chiaburu@sit.fraunhofer.de}} 
    \quad Franz Motzkus$^{3, 4}$
    \quad Frank Haußer$^{5}$\thanks{Equal advising.} 
    \quad Felix Bießmann$^{5, 6}$\footnotemark[2]%
    \\[0.3em]
    \small
    \begin{tabular}{@{}l@{}}
        $^{1}$Fraunhofer SIT \quad 
        $^{2}$ATHENE National Research Center for Applied Cybersecurity \quad
        $^{3}$AUMOVIO \\
        $^{4}$University of Bamberg \quad
        $^{5}$Berlin University of Applied Sciences \quad 
        $^{6}$Einstein Center Digital Future
    \end{tabular}
}

\iclrfinalcopy 
\begin{document}

\maketitle

\begin{abstract}
Vision Sparse Autoencoders (SAEs) have become a popular tool in Mechanistic Interpretability due to their presumed ability to disentangle complex features learned by a model into monosemantic concepts. Despite their growing popularity, evaluating their interpretability remains an active topic of research. The bedrock motivating the adoption of SAEs is the Linear Representation Hypothesis (LRH), which claims that polysemantic features can be projected onto a (near) orthogonal basis of sparse, human-understandable representations. Yet, most current frameworks evaluate proxies such as the sparsity of SAE features or the coherence of the inferred dictionary, implicitly assuming that these reflect alignment with human perception. In this paper, we provide empirical evidence that measuring the interpretability of SAE concepts is more difficult than these proxies suggest. To this end, we adapt the Autointerpretability Score (AIS) - previously shown to align with human judgments in Natural Language Processing - to vision tasks and validate our approach in a dedicated user study. We evaluate SAE concept quality using both standard metrics and our adapted AIS. We find that established interpretability metrics for SAEs correlate neither with one another nor with AIS, indicating that no single reference-free metric, whether grounded in the LRH or not, is sufficient for verifying the interpretability of vision SAEs. We argue these findings support recent calls for more verifiable, ground-truth-anchored design and evaluation of explanation methods. All code and experiments can be found in our anonymized repository: \href{https://anonymous.4open.science/r/XAI_SAE-0222}{https://anonymous.4open.science/r/XAI\_SAE-0222}.
\end{abstract}

\section{Introduction}\label{sec:intro}

The field of Mechanistic Interpretability (MI) aims to reverse-engineer the computational sub-modules of Deep Neural Networks (DNNs) into human-understandable algorithms \citep{olah2020zoom}. Central to this endeavor is the \textit{Linear Representation Hypothesis} (LRH) \citep{park2023linear}, which posits that high-level, human-interpretable concepts are encoded as linear combinations of neurons within the model's activation space. Under this hypothesis, disentangling these features becomes a primary objective for understanding model behavior \citep{bereska2024mechanistic}.

However, a significant roadblock to this understanding is the phenomenon of \textit{polysemanticity}, where individual neurons activate in multiple, semantically distinct contexts. Polysemanticity hinders the identification of concise, univocal explanations for network internals \citep{sae_lm}. Tightly related to this phenomenon is the \textit{superposition} principle \citep{elhage2022superposition}, stating that DNNs represent more features than they have physical neurons by assigning them to an overcomplete set of non-orthogonal directions in the high-dimensional activation space. While efficient for compression, superposition renders the native basis of the network fundamentally uninterpretable.

To resolve this, \textit{Sparse Autoencoders} (SAEs) have recently emerged as a scalable unsupervised method to recover these "ghost" features \citep{bricken2023monosemantic}. An SAE comprises an encoder that projects dense model activations into a higher-dimensional, sparse latent space, and a decoder that reconstructs the original signal. By jointly minimizing reconstruction error and enforcing sparsity, 
SAEs force the model to unpack superimposed features into distinct, monosemantic directions. 

Results on how interpretable SAE features actually are remain, yet, inconclusive. While some studies \citep{sae_lm, sae_protein, million_features} present evidence that features learned by SAEs are more interpretable than native neurons, other authors \citep{interpret_illusions, sae_sanity, projecting_assumptions, same_data_diff_features, sparse_but_wrong} highlight their instability and fragility. 
Moreover, it is still debatable, what metrics and tools should be used to verify the interpretability of the SAE concepts and whether they genuinely align with human perception. In vision tasks, SAEs are very often evaluated only in terms of reconstruction error or the structure of the inferred dictionary, e.g. sparsity of the concept vector and orthogonality of the decoder matrix, omitting user studies to check whether the concepts are human-understandable. Some exceptions include e.g. \cite{sae_monosemantic}, which validate their \textit{Monosemanticity Score} (MS) in a human-in-the-loop (HIL) experiment. In Natural Language Processing (NLP), the community has devised the so-called \textit{Autointerpretability Score} (AIS), which is meant to simulate the human explainer and was shown to correlate with human explanations \citep{ais}. To the best of our knowledge, the literature on vision SAEs was lacking to this day such a measure of interpretability. \autoref{sec:extended_rel_work} describes other established frameworks for automatically describing neurons in vision models, e.g. CLIP-Dissect \citep{clip_dissect} or MILAN \citep{milan}. Yet, they either require a set of pre-annotated concepts or generate unlocalized broad explanations.


Our contributions in this work are as follows:
\begin{enumerate}[nolistsep,leftmargin=.3in]
    \item[\textbf{1)}] By leveraging recent advancements in Large Language Models (LLMs), we propose a recipe for adapting the AIS from NLP to vision SAEs and validate it in a dedicated user survey.
    \item[\textbf{2)}] We show through extensive experiments that standard interpretability metrics for SAEs (whether derived from the LRH, describing the dictionary structure or specifically designed to measure concept disentanglement outside the dictionary structure) do not correlate. Thereby, we argue that concept alignment with human interpretation is a multi-faceted problem, which cannot be fully described by any single metric.
\end{enumerate}

\section{Related Work}\label{sec:rel_work}

\paragraph{Vision SAEs and their Evaluation.}

After wide adoption in NLP for explaining intermediate activations of LLMs, SAEs have gained increasing attention in Computer Vision (CV) as well.
While early approaches applied standard "Vanilla" ReLU-based $L_1$-constrained SAEs \citep{bricken2023monosemantic} to vision transformer embeddings \citep{sae_rigorous,probing_sae}, recent work has explored more specialized architectures for enforcing sparsity and concept purity. Just to name a few examples: JumpReLU SAEs \citep{jump_relu}, TopK SAEs \citep{topk_sae, batchtopk_sae}, Matryoshka \citep{matryoshka}, Matching Pursuit (MP-SAE) \citep{matching_pursuit}, Archetypal (A-SAE) \citep{archetypal_sae}, Universal SAE \citep{universal_sae}. In their original investigations, these SAE classes were trained to reconstruct "global" image features, e.g. the CLS token in standard transformer architectures such as CLIP \citep{clip}.  
Other authors \citep{patch_sae, sae_rigorous} target more localized concept learning and propose training SAEs on patch-level features, which any modern vision encoder extracts when tokenizing the image. 

As far as SAE validation is concerned, verifying interpretability and identifying semantics within SAE neurons remains challenging.
Although meant as an MI tool, SAEs are oftentimes evaluated solely on indicators only indirectly related to interpretability, such as the reconstruction error, 
the sparsity of the concepts or the orthogonality of the inferred dictionary. While explanation simplicity is frequently cited as a desideratum \citep{nauta_co12}, trading completeness for sparsity does not strike the right compromise and is known to erode users' trust in the model's prediction just as much as an overly complex explanation \citep{haufe_paradigm}. 

Alternative proxies from the literature include: (i) approaches measuring the effect of SAEs on the downstream task performance \citep{sae_rigorous, cafe, patch_sae} or (ii) similarity-based approaches, e.g. in the explained input space, such as the \textit{Monosemanticity Score} \citep{sae_monosemantic} or w.r.t. to an external concept ontology such as WordNet \citep{probing_sae}. In \citep{archetypal_sae}, the authors look at various similarities in the dictionary learning process, e.g. the \textit{OOD score} (similarity between the dictionary atoms and the image embeddings on which it was learned), \textit{Coherence} (self-similarity of the dictionary itself) or \textit{Connectivity} (self-similarity of the SAE features). We will return to some of these metrics later in \autoref{sec:results}.

Recently, more and more researchers in the XAI (Explainable AI) community are advocating for more formalization in the design of the explainers and supervision on ground truth explanations \citep{haufe_formal, haufe_paradigm, luca_manifesto}. Fittingly, recent contributions in the SAE literature pick up this direction. For instance,
\cite{concept_sae} utilize a pipeline with ground-truth Vision-Language Models (VLMs) and segmentation masks to evaluate how faithfully SAEs reconstruct specific visual concepts. \cite{sae_concept_annot} evaluate their interpretability against annotated concepts on CUB-200 \citep{cub} and COCO \citep{coco}. \cite{concept_sae} propose Concept-SAE, which aligns the training process with user-defined concepts.


\paragraph{Autointerpretability for SAEs in Sequence Modelling. }

The methodology of automated interpretability scoring was first introduced in the context of LLMs.
\citep{sae_lm} established the baseline for this approach using the AIS, originally proposed by \cite{ais}. By training SAEs on Pythia-70M \citep{pythia} embeddings from OpenWebText \citep{openwebtext}, they demonstrated that SAE features exhibit reduced polysemanticity compared to linear baselines like PCA and ICA (Independent Component Analysis), particularly in earlier model layers. Building on this, \cite{million_features} scaled the analysis to millions of features across varied SAE architectures and activation functions trained on RedPajama-v2 \citep{redpajama}. They also proposed more compute-efficient modifications to the AIS. 

The success of SAEs in NLP has extended to other sequence-based domains, such as Protein Language Models (PLMs). \cite{sae_protein} applied TopK SAEs \citep{topk_sae} to the protein-level and amino-acid representations of an ESM2 model \citep{esm}, treating protein sequences analogously to text. They validated the biological relevance of the resulting sparse features not only via AIS, but also through post-hoc Gene Ontology (GO) analysis, revealing strong associations between specific neurons and functional annotations in the UniProt database \citep{uniprot}.


\section{Dictionary Learning with SAEs}\label{sec:saes}

In this section we introduce the mathematical objects we will be working with throughout this paper and discuss the definition and assumptions of the LRH. In \autoref{sec:extended_rel_work} we discuss more related work on the origins of sparse coding and dictionary learning.

\subsection{Notation}

\paragraph{Patch-level Representations. }

Let $x \in \mathbb{R}^{d}$ denote a flattened RGB-image, where $d = \text{width} \cdot \text{height} \cdot 3$. Across all configurations, we resize our images into square $224 \times 224 \times 3$ formats, hence $d$ is constant. We denote the image dataset of size $n$ by $X \in \mathbb{R}^{n \times d}$, where the row $x_i^T$ is the $i$-th flattened image, $\forall 1 \leq i \leq n$. Throughout this paper, vectors are per default column vectors, unless they are transposed; then, they are meant as row vectors.

We extract patch-wise visual features/embeddings from $X$ with an embedder $\varepsilon: \mathbb{R}^d \rightarrow \mathbb{R}^{s \times p}$, where $s$ is the number of patches that the embedder splits the image into and $p$ is the dimension of the embedding space (which the SAE will be trained to reconstruct). We describe the patch embeddings block $\varepsilon(x_i) \in \mathbb{R}^{s \times p}$ as $E_i$, which collects all $s$ patch embeddings of image $x_i$. Stacking all $n$ $E_i$-blocks together builds the full patch embeddings matrix $E \in \mathbb{R}^{(ns) \times p}$, where each row vector $e_j^T \in \mathbb{R}^p$ is one patch embedding, $\forall 1 \leq j \leq ns$.

In our pipeline, the SAE maps each $e_j$ into a (higher-dimensional) latent space $\mathbb{R}^{m}$, to extract sparse representations. Typically, we have $m > p$, which presumes an \textit{overcomplete} set of sparse features. We note that, similar to \citep{patch_sae}, we will train our SAEs on patch-level embeddings, not on global image-level representations, as is usually the case in the literature. This is necessary for the computation of the AIS later. Image-level features will, however, still play a role in other metrics, as will be discussed in the following sections.

The SAE architecture consists of an encoder and a decoder. The encoder computes the sparse patch-level codes/concepts\footnote{We use interchangeably the terms \textit{SAE features}, \textit{concepts}, \textit{neurons} and \textit{codes}.} $z_j \in \mathbb{R}^m$ using a learned weight matrix $W_{enc} \in \mathbb{R}^{m \times p}$, an encoder bias $b_{enc} \in \mathbb{R}^{m}$, a pre-decoding bias $b_{pre} \in \mathbb{R}^{p}$ (usually chosen as the geometric median of the training set), and a non-linear activation function $\sigma$:
\begin{equation}\label{eq:sae_enc}
    z_j = \sigma(W_{enc}(e_j - b_{pre}) + b_{enc})
\end{equation}
We denote by $Z \in \mathbb{R}^{(ns) \times m}$ the matrix that collects all sparse codes mapped from $E$. In this context, $Z_i \in \mathbb{R}^{s \times m}$ will be the block of codes corresponding to $E_i$. Implicitly, each column in $Z$ contains all activations for each SAE neuron; we denote this as the activation $a^{(k)} \in \mathbb{R}^{ns}$ for the $k$-th SAE neuron, $\forall 1 \leq k \leq m$. We warn the reader that various metrics throughout our paper will require (conceptually) either the columns or the rows of particular collections, such as $Z$, hence the targeted different symbols to avoid confusion. 

The decoder subsequently reconstructs the original embedding vector to produce $\hat{e}_j \in \mathbb{R}^{p}$ using a \textit{dictionary matrix} $W_{dec} \in \mathbb{R}^{p \times m}$:
\begin{equation}\label{eq:sae_dec}
    e_j \approx \hat{e}_j = W_{dec}z_j + b_{pre}
\end{equation}
In the SAE literature, the transposed decoder matrix $W_{dec}^T$ is usually referred to as the \textit{dictionary} $D \in \mathbb{R}^{m \times p}$, where each row $d_k^T \in \mathbb{R}^{p}$ is known as a \textit{dictionary atom}.

\paragraph{Image-level Representations. } 

As mentioned previously, certain calculations will require global image-level embeddings, codes or activations, not patch-level ones. Therefore, we introduce the following additional objects:

\begin{itemize}[nolistsep,leftmargin=.3in]
    \item $\tilde{e}_i \in \mathbb{R}^p$ - the image-level embedding for $x_i$. It is either delivered automatically by some embedders in the form of the CLS token or we compute it ourselves as $\tilde{e}_i = \frac{1}{s} \sum_{j = (i-1)s+1}^{is} e_j$. Accordingly, the matrix collecting all these image embeddings is denoted by $\tilde{E} \in \mathbb{R}^{n \times p}$.

    \item $\tilde{z}_i \in \mathbb{R}^m$ - the image-level codes for $x_i$. We compute them by max-pooling the corresponding patch features: $\tilde{z}_{i,k} = \underset{(i-1)s+1 \leq j \leq is}{\max} z_{j,k}$. They will be stacked in $\tilde{Z} \in \mathbb{R}^{n \times m}$.

    \item $\tilde{a}^{(k)} \in \mathbb{R}^n$ - the similarly pooled image activations in the $k$-th SAE neuron ($k$-th column in $\tilde{Z}$).
\end{itemize}

\subsection{Linear Representation Hypothesis}\label{subsec:lrh}



SAEs have increasingly drawn attention in recent years, especially as MI has gained contour as a subfield within XAI. They promise to offer a straightforward post-hoc solution to the polysemanticity problem addressed in \autoref{sec:intro}. The viability of SAEs as a solution to this problem is primarily motivated by the LRH, also known as the \textit{superposition hypothesis} \citep{bricken2023monosemantic, elhage2022superposition}. In this work, we will lean on the LRH formulation in \citep{matching_pursuit}: \textit{"[...] high-dimensional neural representations can be decomposed as superpositions over a large set of approximately orthogonal directions, each aligned with human-interpretable concepts."}

In other words, assuming such a near-orthogonal human-interpretable basis exists, the dictionary $D$ learned by an SAE should converge towards it. Without loss of generality, we leave out the constant $b_{pre}$ and patch index $j$ from \autoref{eq:sae_dec} and have:
    $e \approx \hat{e} = W_{dec}z = D^T z = \sum_{k=1}^m z_k \cdot d_k$, with $z_k \in \mathbb{R}$,
under the following assumptions:
\begin{enumerate}[nolistsep,leftmargin=.3in]
    \item[\textbf{1)}] Overcompleteness: $m \gg p$
    \item[\textbf{2)}] Near-Orthogonal Dictionary: 
        $\underset{k \neq l}{\max} \left| d_k^\top d_l \right| \leq \delta, \quad \|d_k\|_2=1, \forall k$
    \item[\textbf{3)}] $K$-Sparsity: $\|z\|_0^{} \leq K \ll m.$
\end{enumerate}
The operator $\left\| \cdot \right\|_0^{}$ is the $L_0$ pseudo-norm (number of non-zero entries). The second condition corresponds to a low-coherence dictionary.
While conditions \textbf{1-3} are commonly optimized for as structural proxies for a sparse linear decomposition of the entangled concepts in $e \in \mathbb{R}^p$, e.g. in \citep{bricken2023monosemantic, topk_sae, archetypal_sae, batchtopk_sae, matryoshka}, satisfying them offers no theoretical guarantee that the resulting decomposition aligns with human-interpretable concepts, as LRH requires. We argue that the three conditions are at most necessary but not sufficient for LRH: alignment with human perception is only implicitly assumed, not established, by construction. 
As we will show, measuring this alignment is more difficult and metrics specifically designed to quantify it fail to correlate with proxies such as reconstruction error or orthogonality of the dictionary.

Other authors \citep{projecting_assumptions, matching_pursuit, relational_composition, park2024, csordas_recurrent, engels2024decomposing, engels2025onedim} 
also discuss cases where the ground truth structure of concepts is such that concepts are not linearly separable, heterogeneous (different concepts reside in subspaces of different dimensions) or interact with each other. \cite{sae_manifolds} argue in favor of analyzing \textit{concept manifolds} instead of individual directions in the SAE activation space. On a broader level, recent work on validating SAEs \citep{interpret_illusions, sae_sanity, projecting_assumptions, same_data_diff_features, sparse_but_wrong} pinpoints the fragility of the concepts they learn, their lack of robustness and unreliability, which naturally complements the growing body of literature criticizing the instability and foundational gaps in current explanation methods, e.g. \cite{robustness_interp, salcritique1, salcritique2, salcritique3, salcritique4, jyoti2022robustnessexplanationsdeepneural}. 

\section{Methods}\label{sec:methods}

This section describes our implementation of the AIS for Vision SAEs, as well as the datasets and models used in our experiments. More SAE training details can be found in \autoref{sec:sae_training} and the other standard SAE evaluation metrics we took into consideration (reconstruction $R^2$, $L_0$-sparsity, OOD Score, Coherence, Connectivity and MS) are described in \autoref{sec:other_metrics}.

\subsection{AIS Procedure}

\begin{wrapfigure}{R}{0.5\textwidth}
    \vspace{-10pt}
    \centering
    \includegraphics[width=1\linewidth]{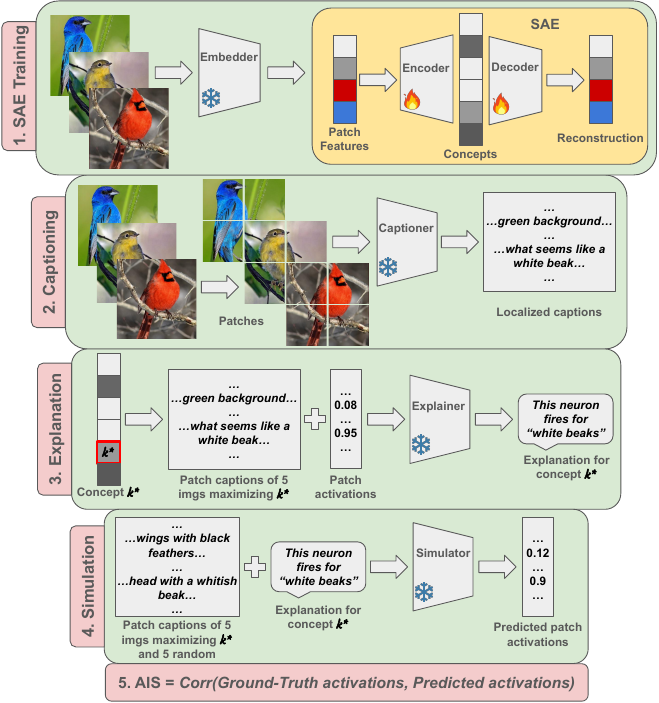}
    \caption{\textbf{AIS Pipeline: (1)} Train SAE on patch embeddings. \textbf{(2)} 
    Split images into a grid and caption every grid patch.
    \textbf{(3)} For a given neuron $k^*$, feed pairs of patch captions and activations from 5 high activating images into the LLM-explainer and prompt it to infer the common concept. \textbf{(4)} In reverse, show the LLM-simulator pairs of patch captions along with the previously inferred explanation (5 from high-activating images, 5 from random images) and prompt it to predict the neuron activations per patch. \textbf{(5)} Compute the AIS of the neuron $k^*$ as the Pearson correlation between predicted and true patch activations.}
    \label{fig:ais_pipeline}
    \vspace{-10pt}
\end{wrapfigure}

\autoref{fig:ais_pipeline} outlines our proposed adaptation scheme of AIS for Vision SAEs. 
The \textbf{first step} consists of training an SAE to reconstruct the embeddings $e \approx \hat{e}$. Once fully trained, only the encoder module will be used further to generate the sparse codes $z$.

In a parallel \textbf{second step}, the images are split into 
patches and each grid patch is annotated by a captioning model. This is where the novelty arises when transferring the AIS method from language to vision. When it comes to text, AIS localizes SAE-learned concepts on the token level, depending on which tokens in a body of text activate a specific neuron in the SAE. For images, this operation is not as straightforward. While modern embedders do split up images into "tokens" (patches), most vision SAEs are trained on full image embeddings, which does not match conceptually with the idea of AIS. Hence, we propose working with patch-level embeddings and SAE activations here, to allow similar iterations for the AIS as in the textual scenario. The captions are, therefore, meant to translate the image patches into sentences. 
Note that the captioning grid resolution is a hyperparameter and will highly influence the quality of the downstream explanations. 

The \textbf{third step} is the explanation generation. For a subset $K^*$ of relevant SAE neurons (see \autoref{sec:max_filter}), we select the top 10 images that maximize (by average of patches) each neuron's activation. Then, an explainer (in form of an LLM) is shown the pairs of captions and normalized\footnote{Following \cite{ais}, we normalize activations onto a discrete 0-10 integer interval. Negative values are mapped to 0, the maximum value in that neuron is mapped to 10.} patch activation scores for 5 of those images. The explainer is prompted to infer the common explanation/concept for which that particular neuron fires. An example of an explained neuron can be seen in \autoref{fig:output_ais_explain} in the Appendix.

In the \textbf{fourth step}, another or the same LLM acts as a simulator. Based on the previously inferred explanation, the other top 5 activating images (namely, their patch captions) and 5 other random images (known as the "top-random" approach \citep{sae_lm}), the LLM is prompted to predict the SAE activation score for every patch (see an example in \autoref{fig:output_ais_simulate} in the Appendix).

In the final \textbf{fifth step}, we compute the AIS per explained neuron as the Pearson correlation between the ground truth patch activation scores $A_{\text{truth}}$ and the simulated ones $\hat{A}_{\text{pred}}$. The average of all the neuron-wise AIS values gives the Mean AIS as a global measure of interpretability for the SAE.


\subsection{Data and Experiments}

\begin{wrapfigure}{R}{0.5\textwidth}
    \vspace{-15pt}
    \begin{algorithm}[H]
        \SetAlgoLined
        \DontPrintSemicolon
        \KwIn{Datasets $\mathcal{D}$, Embedders $\mathcal{E}$, SAEs $\mathcal{S}$, Captioners $\mathcal{C}$, Filtered Index Set $K^*$, LLM-Explainer and -Simulator}
        \KwOut{AIS} 
        
        \For{$\textbf{d} \in \mathcal{D}, \varepsilon \in \mathcal{E}, \textbf{s} \in \mathcal{S}, \textbf{c} \in \mathcal{C}, k^* \in K^*$}{
            \tcp{\textbf{Captioning}}
            $I_{\text{top}} \leftarrow$ Top 10 images maximizing activation of neuron $k^*$\;
            $C_{\text{patches}} \leftarrow$ Caption patches in $I_{\text{top}}$ with \textit{\textbf{c}}\;
            
            \tcp{\textbf{LLM Explanation}}
            $P_{\text{exp}} \leftarrow$ Select 5 pairs (Patch Captions, Activations neuron $k^*$) from $I_{\text{top}}$\;
            $Concept_{k^*} \leftarrow$ LLM($P_{\text{exp}}$, \text{"Infer concept"})\;
            
            \tcp{\textbf{LLM Simulation}}
            $P_{\text{sim}} \leftarrow$ Select 10 new pairs (Patch Captions, $Concept_{k^*}$); 5 high activating + 5 random\;
            $\hat{A}_{\text{pred}} \leftarrow$ LLM($P_{\text{sim}}$, \text{"Predict activation"})\;
            $A_{\text{truth}} \leftarrow$ Ground truth activations for $P_{\text{sim}}$\;
            
            \tcp{\textbf{Evaluation}}
            $AIS_{k^*} \leftarrow \text{Correlation}(\hat{A}_{\text{pred}}, A_{\text{truth}})$\;
        }
        Average $AIS$ over $|K^*|$
        \caption{\textbf{AIS Experiments}. $\mathcal{D} =$ \{CUB, ImageNet100, Caltech\}; $\mathcal{E} =$ \{DINO, ViT, SigLIP\}; $\mathcal{S} =$ \{Vanilla, TopK, MP-SAE\}; $\mathcal{C} =$ \{LLaVa\}; LLM = \{Gemma\}.}\label{algo:ais}
    \end{algorithm}
    \vspace{-10pt}
\end{wrapfigure}

The pseudocode in \autoref{algo:ais} gives an overview of our AIS experiments.
We run the experiments described in this paper on the following datasets: CUB-200-2011 \citep{cub}, 5794 samples; a subset of ImageNet \citep{imagenet} with 100 classes containing around 200 samples each - we denote it here as ImageNet100
; and Caltech 101 \citep{caltech}, 4572 samples. For CUB, we applied the standard split that comes with downloading the dataset. As for the other two, we applied a fixed-seed stratified split of 0.5-0.5 into training and validation sets. We extract embeddings from the images in these datasets with various embedders $\varepsilon$, namely: DINOv2 \citep{dino}, ViT \citep{vit} and SigLIP \citep{siglip}. While we conduct experiments on single-object, object-centric datasets, our approach can readily be applied on other vision benchmarks as well.

As noted above, the captioning grid size is a hyperparameter that influences the quality of the explanations. After several iterations, we settled for a grid size of 4 as a middle ground. Hence, we split each image into a $4 \times 4$ grid and  caption each patch using LLaVa 1.5 \citep{llava}. Example captions can be found in the Appendix. We limit the maximum number of generated caption tokens to 100 and prompt the model as follows: \\
\textit{
Directly describe the texture, shapes, and colors visible in this close-up image. Be concise and focus only on visual details.
}

As explainer and simulator we tested various modern open-weight LLMs that can be run on-premise, namely from the Gemma 3 \citep{gemma3} and Gemma 4 \citep{gemma4} families. We decided to use the same model as both explainer and simulator, but note that one can delegate two different models for the two tasks (see e.g. \cite{sae_lm}).

\section{Results and Discussion}\label{sec:results}

In this section, we present our findings and discuss their possible interpretations and implications.

\paragraph{1) LLM Explanations Align with Human Judgment. }


Previous research in NLP has shown through HIL experiments that LLM explanations align well with human judgments \citep{ais}. To the best of our knowledge, a similar investigation was missing for vision tasks. To this end, we designed a user survey, in order to verify the soundness of our vision AIS procedure. The design of the experiment is described in \autoref{sec:user_study}.

\begin{wrapfigure}{r}{0.5\textwidth}
    \vspace{-13pt}
    \centering
    \begin{minipage}{\linewidth}
        \centering
        \captionsetup{type=table}
        \caption{\textbf{LLM vs Human Explanations.} Comparison of AIS scores on two MP SAE neurons for Gemma and human users. AIS LLM and AIS Users (averaged per user) are both computed on the $[0, 2]$ range of scores.}
        \label{tab:user_study}
        \resizebox{\linewidth}{!}{%
        \begin{tabular}{l|cc}
        \toprule
        \textbf{Neuron} & \textbf{AIS LLM} & \textbf{Mean AIS Users} \\
        \hline
        2174 ("branches") & \textbf{0.6644} & 0.5082 $\pm$ 0.1480 \\
        9948 ("water")    & \textbf{0.8800} & 0.6263 $\pm$ 0.3091 \\
        \bottomrule
        \end{tabular}%
        }
        \vspace{1em}

        \begin{minipage}{0.49\linewidth}
            \centering
            \includegraphics[width=\linewidth]{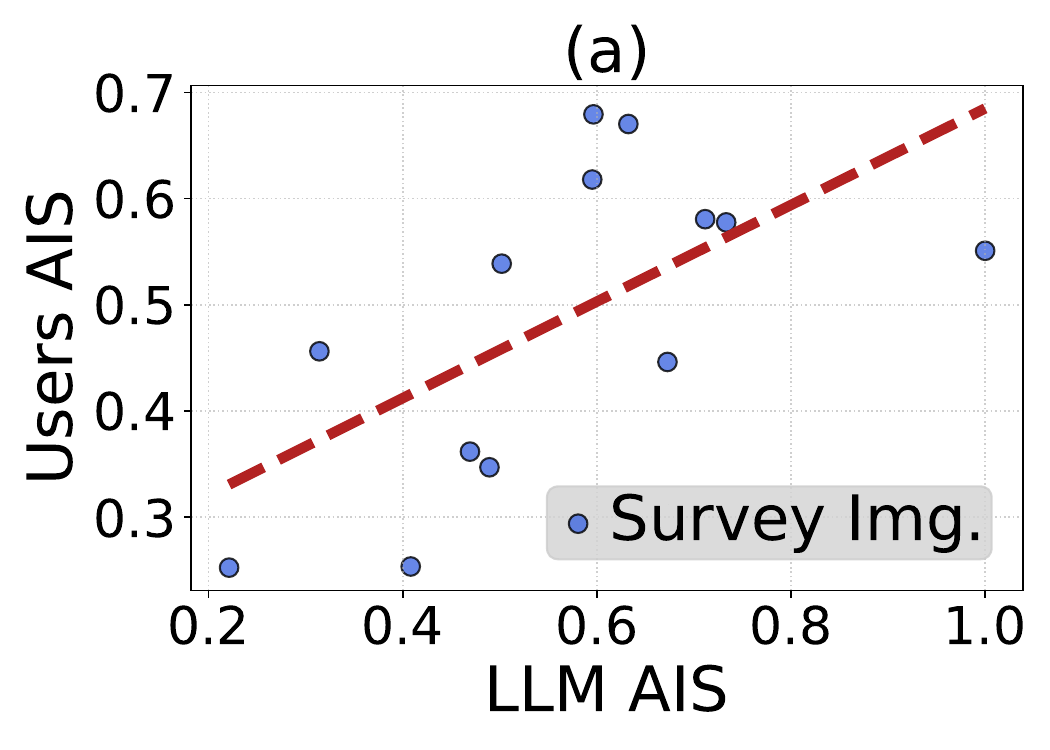}
        \end{minipage}\hfill
        \begin{minipage}{0.49\linewidth}
            \centering
            \includegraphics[width=\linewidth]{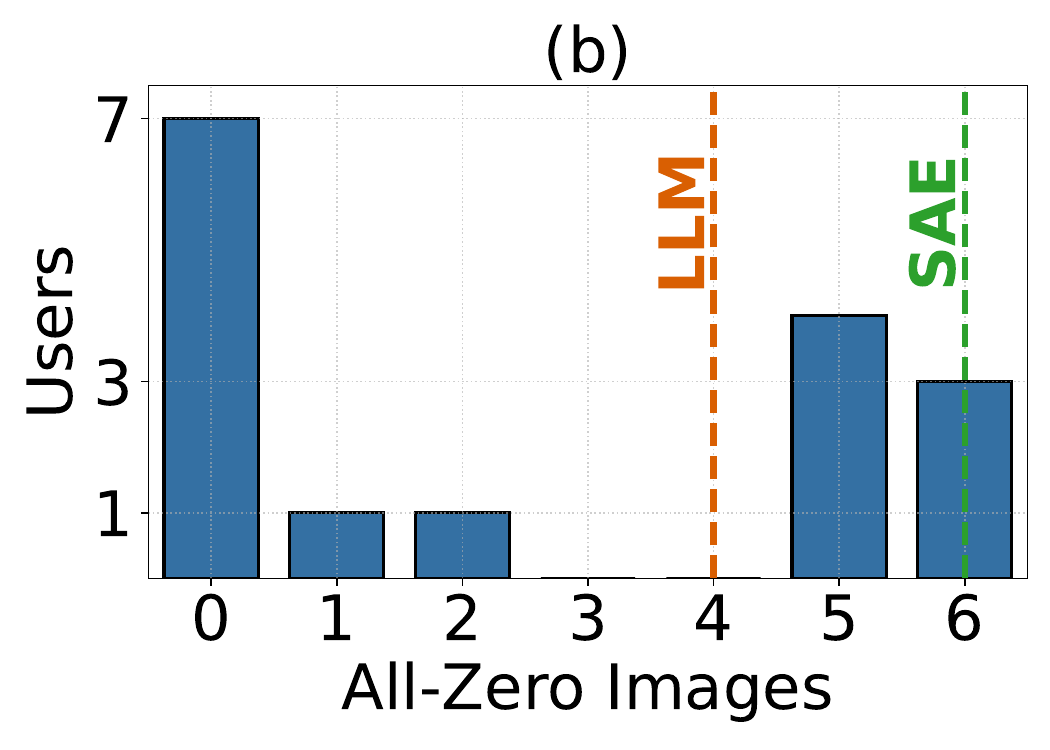}
        \end{minipage}
        \captionsetup{type=figure}
        \caption{\textbf{(a) LLM-AIS correlates with Human-AIS.} Each point represents a non-degenerate survey image (14 in total). All AIS values are computed on the $[0, 2]$ score range. \textbf{(b) 7/16 users outperform the LLM in recognizing trivial all-zero cases.}}
        \label{fig:ais_llm_users}
    \end{minipage}
    \vspace{-20pt}
\end{wrapfigure}

Plot (a) in \autoref{fig:ais_llm_users} shows a positive correlation - yet rather imprecisely estimated given the small sample size - between the image-level AIS values achieved by Gemma and survey participants: coefficient $\textbf{r = 0.6202}$, at a p-value of \textbf{0.0237} and a 95\% confidence interval \textbf{[0.133, 0.866]} (via Fisher's z-transformation). This suggests that the caption-based adaption of AIS is able to describe SAE concept quality similar to how human explainers perceive it. Out of the 20 simulation images used in the survey we excluded 6 degenerate cases (all-zero ground-truth SAE scores) from plot (a) and analyzed them separately in plot (b). The LLM correctly identified 4 of these cases, while 7 out of 16 users also managed to identify all or almost all of them. Upon manual inspection, we confirmed that these users formulated more detailed descriptions of the concepts in the first task than the rest - \autoref{fig:user_explanations}. Qualitatively, this gives reassurance that i) the LLM simulator does not tend to hallucinate activation scores where the concept is absent and ii) users are also able to correctly annotate trivial cases, provided they find a concept specific enough in the explanation task. For a second part of the discussion regarding degenerate cases, see our ablation study in \autoref{sec:ablation}.

\autoref{tab:user_study} compares the individual AIS on the two neurons for the Gemma Explainer/Simulator against the users. The results indicate that our automated LLM pipeline outperforms human users in extracting explanations and simulating activations, even when the image material is translated into captions and the activation range is more fine-grained. This, along with the random shuffling test in \autoref{tab:baselines}, is another proof of concept in favor of the caption-based adaptation of the AIS procedure.

After inspecting the users' explanations for both neurons (\autoref{fig:user_explanations}), we learned that the simulation task was notably more challenging, even though most users successfully identified "branches" and "water" as concepts, similar to Gemma's explanations. This is also indicated by the fairly high disagreement rate between users (in form of the standard deviations in \autoref{tab:user_study}). This insight can be used to further optimize the efficiency of the AIS pipeline, e.g. by using a lower-scale LLM for explanation than for simulation (see \cite{sae_lm}).

\begin{table*}[htbp]
\centering
\caption{\textbf{High reconstruction - low interpretability.} Best results are marked in \textcolor{ForestGreen}{\textbf{green}}, worst results in \textcolor{red}{\textbf{red}}. More statistics accompanying the AIS values are collected in \autoref{tab:ais_extra_metrics} from the Appendix. In particular, note that Sparsity, as defined in \autoref{eq:sparsity} and in the \textit{overcomplete} library \citep{overcomplete}, actually measures density (meaning a higher value is denser). }
\label{tab:sae_metrics}
\resizebox{\textwidth}{!}{
\begin{tabular}{lllcccccc||c}
\hline
\textbf{SAE} & \textbf{Embedder} & \textbf{Dataset} & \textbf{Sparsity} & \textbf{$R^2 (\uparrow)$} & \textbf{OOD Score $(\downarrow)$} & \textbf{Coherence $(\downarrow)$} & \textbf{Connectivity $(\downarrow)$} & \textbf{Mean MS $(\uparrow)$} & \textbf{Mean AIS $(\uparrow)$} \\
\hline
Vanilla & DINO   & CUB200      & 0.5012 & 0.9706 & 0.7898 & 0.3146 & \color{red}{\textbf{1.0000}} & 0.0500 & 0.3493 \\
        &        & ImageNet100 & 0.5007 & 0.9254 & 0.7669 & 0.3146 & \color{red}{\textbf{1.0000}} & 0.0059 & 0.2801 \\
        &        & Caltech     & 0.5009 & 0.9511 & 0.7781 & 0.3146 & \color{red}{\textbf{1.0000}} & 0.0124 & 0.2049 \\
\cline{2-10}
        & ViT    & CUB200      & 0.4990 & 0.9629 & 0.8528 & \color{ForestGreen}{\textbf{0.2111}} & \color{red}{\textbf{1.0000}} & \color{ForestGreen}{\textbf{0.1060}} & 0.1621 \\
        &        & ImageNet100 & 0.4985 & 0.9296 & 0.8352 & \color{ForestGreen}{\textbf{0.2111}} & \color{red}{\textbf{1.0000}} & 0.0808 & 0.2105 \\
        &        & Caltech     & 0.4980 & 0.9322 & 0.8444 & \color{ForestGreen}{\textbf{0.2111}} & \color{red}{\textbf{1.0000}} & 0.0825 & 0.1510 \\
\cline{2-10}
        & SigLIP & CUB200      & 0.5019 & \color{ForestGreen}{\textbf{0.9901}} & \color{red}{\textbf{0.8662}} & \color{ForestGreen}{\textbf{0.2111}} & \color{red}{\textbf{1.0000}} & 0.0146 & 0.3518 \\
        &        & ImageNet100 & 0.5016 & 0.9777 & 0.8492 & \color{ForestGreen}{\textbf{0.2111}} & \color{red}{\textbf{1.0000}} & 0.0033 & 0.3545 \\
        &        & Caltech     & 0.5012 & 0.9736 & 0.8549 & \color{ForestGreen}{\textbf{0.2111}} & \color{red}{\textbf{1.0000}} & 0.0068 & 0.2522 \\
\hline\hline
TopK   & DINO   & CUB200      & 0.1965 & 0.9596 & 0.6541 & 0.8750 & \color{red}{\textbf{1.0000}} & 0.0221 & 0.1619 \\
        &        & ImageNet100 & 0.1999 & 0.9180 & 0.6529 & 0.9509 & \color{red}{\textbf{1.0000}} & 0.0026 & 0.0820 \\
        &        & Caltech     & 0.1970 & 0.9135 & 0.6785 & 0.7463 & \color{red}{\textbf{1.0000}} & 0.0051 & 0.0636 \\
\cline{2-10}
        & ViT    & CUB200      & 0.1993 & 0.9521 & 0.7201 & \color{red}{\textbf{1.0000}} & 0.9972 & 0.0580 & \color{red}{\textbf{0.0226}} \\
        &        & ImageNet100 & 0.1995 & 0.9188 & 0.7303 & 0.9221 & \color{red}{\textbf{1.0000}} & 0.0489 & 0.0505 \\
        &        & Caltech     & 0.1993 & 0.9034 & 0.7293 & 0.9022 & \color{red}{\textbf{1.0000}} & 0.0450 & 0.0477 \\
\cline{2-10}
        & SigLIP & CUB200      & 0.1980 & 0.9894 & 0.6320 & \color{red}{\textbf{1.0000}} & \color{ForestGreen}{\textbf{0.5584}} & 0.0089 & 0.1458 \\
        &        & ImageNet100 & 0.1977 & 0.9728 & 0.6879 & \color{red}{\textbf{1.0000}} & 0.8312 & 0.0023 & 0.0587 \\
        &        & Caltech     & 0.1977 & 0.9660 & 0.7226 & \color{red}{\textbf{1.0000}} & 0.9738 & 0.0046 & 0.0779 \\
\hline\hline
MP      & DINO   & CUB200      & 0.0081 & 0.9858 & 0.6836 & 0.9600 & \color{red}{\textbf{1.0000}} & 0.0090 & 0.3476 \\
        &        & ImageNet100 & 0.0081 & 0.9792 & \color{ForestGreen}{\textbf{0.5944}} & 0.9530 & \color{red}{\textbf{1.0000}} & 0.0010 & \color{ForestGreen}{\textbf{0.4332}} \\
        &        & Caltech     & 0.0081 & 0.9798 & 0.6586 & 0.9167 & \color{red}{\textbf{1.0000}} & 0.0021 & 0.3763 \\
\cline{2-10}
        & ViT    & CUB200      & 0.0081 & 0.9265 & 0.7573 & 0.9531 & \color{red}{\textbf{1.0000}} & 0.0152 & 0.2817 \\
        &        & ImageNet100 & 0.0081 & \color{red}{\textbf{0.8909}} & 0.6894 & 0.9280 & \color{red}{\textbf{1.0000}} & 0.0101 & 0.4222 \\
        &        & Caltech     & 0.0081 & 0.9047 & 0.7310 & 0.8399 & \color{red}{\textbf{1.0000}} & 0.0106 & 0.3834 \\
\cline{2-10}
        & SigLIP & CUB200      & 0.0081 & 0.9749 & 0.7259 & 0.9824 & \color{red}{\textbf{1.0000}} & 0.0028 & 0.2750 \\
        &        & ImageNet100 & 0.0081 & 0.9564 & 0.6629 & 0.9858 & \color{red}{\textbf{1.0000}} & \color{red}{\textbf{0.0004}} & 0.3213 \\
        &        & Caltech     & 0.0081 & 0.9494 & 0.7019 & 0.9914 & \color{red}{\textbf{1.0000}} & 0.0011 & 0.2764 \\
\bottomrule
\end{tabular}%
} 
\end{table*}

\paragraph{2) High Reconstruction - Low Interpretability. }
\autoref{tab:sae_metrics} confirms a trade-off well documented in the SAE and XAI literature \citep{Herm_2023, revisiting_perf_exp_tradeoff}: A high reconstruction potential does not guarantee the SAE is also interpretable. While all SAE configurations reach a high $R^2$ score on reconstructing the embeddings, all other interpretability metrics indicate that the SAE features are still rather opaque. In particular, all configurations achieve very high OOD, Coherence and Connectivity scores; the only exception here is the Coherence below 0.5 for the Vanilla SAEs. Also immediately visible is the maximum Connectivity for almost all the configurations, giving a first piece of evidence that the SAE concepts remain polysemantic. 
On the opposite side, AIS and MS are both low and extremely low, respectively, none of them surpassing 0.5. When compared against baselines from the literature: Our vision SAEs trained here on patch-level embeddings achieve a much lower MS range - $[4 \times 10^{-4}, 1.06 \times 10^{-1}]$ - than their counterparts trained on image-level embeddings from the original paper introducing the MS score \citep{sae_monosemantic} - roughly $[0.1, 0.6]$ for a similar SAE expansion factor as ours on the final embeddings. As for the AIS: Our vision SAEs compare favorably against SAEs explaining LLMs \citep{sae_lm} and are surpassed by SAEs explaining PLMs \citep{sae_protein} (also refer to \autoref{tab:baselines}). In terms of sparsity-interpretability trade-off: MP-SAEs fall out as achieving comparable results as the other SAEs, while trained on a much lower sparsity regime.
This highlights again the argument introduced in \autoref{sec:rel_work}: The balance between explanation simplicity (approximated here in the form of sparsity of SAE features) and completeness is difficult to find and describe accurately.

\paragraph{3) Interpretability Metrics Do Not Correlate as Expected. }

\begin{wrapfigure}{R}{0.5\textwidth}
    \vspace{-20pt}
    \centering
    \includegraphics[width=1\linewidth]{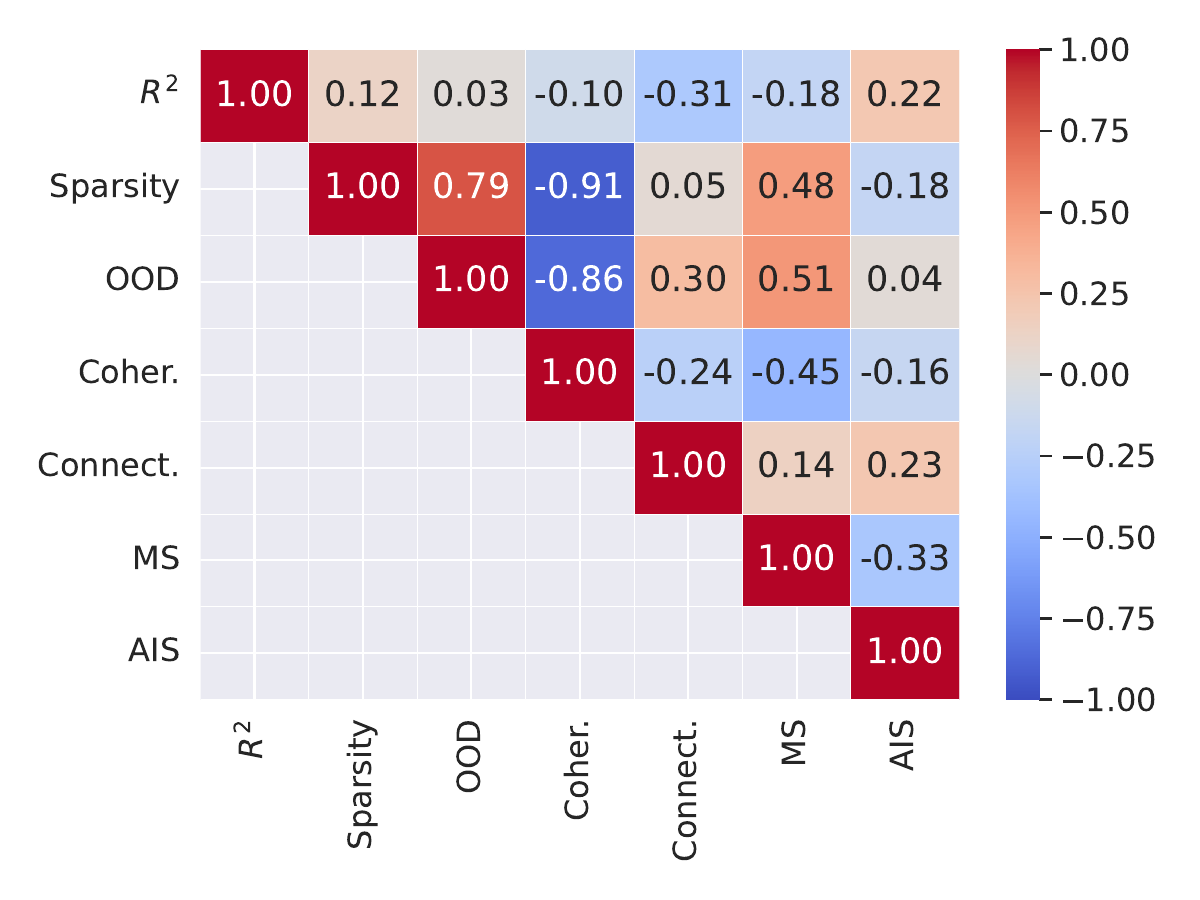}
    \vspace{-2.5em}
    \caption{\textbf{Correlation matrix on the SAE metrics.} Except (Sparsity - OOD Score), none of the other pairs correlate as expected.}
    \label{fig:corr_mat}
    \vspace{-10pt}
\end{wrapfigure}

\autoref{fig:corr_mat} depicts the correlation matrix between the metrics in \autoref{tab:sae_metrics}. What we learn from it is that standard SAE evaluation metrics (for reconstruction and interpretability) do not correlate more than moderately, except for three cases: (Sparsity - OOD Score), (Sparsity - Coherence) and (OOD Score - Coherence). While the threshold for a "high correlation" is subjective and task-dependent, we consider here $\pm0.6$ to be suitable \citep{guide_corr}.

Note that while Sparsity and Coherence, which are central pieces of the LRH (\autoref{subsec:lrh}), do correlate strongly, the negative relationship of this correlation contradicts the requirements of the LRH, namely sparse near-orthogonal dictionaries. According to these assumptions, one would expect to observe sparse features and low-coherence dictionaries together. Similarly for OOD Score vs Coherence, where a positive correlation was to be expected. Only Sparsity and OOD Score exhibit an expected high positive correlation.
Of particular interest are MS and AIS, which were designed to measure interpretability from outside the dictionary structure. MS and AIS do not appear to correlate w.r.t. each other; w.r.t. the other metrics, none of them are highly correlated.

\paragraph{4) Captioning and Smaller LLMs Integrate Well in AIS. }

We experimented with various Gemma models as explainers and simulators, on a fixed configuration Vanilla SAE + DINO + CUB (\autoref{tab:baselines}). While some lower scale Gemma versions struggled as explainers and simulators, some versions of as few as 4B parameters achieved similar AIS values as models of three or six times their size.
We run Gemma inference on-premise to generate the explanations and simulate the activations for all versions up to 12B parameters. Gemma4-26b is called in the cloud via Google's API in the Google AI Studio.
Note that our AIS results with much smaller LLM explainers/simulators are comparable to similar AIS experiments from the NLP and PLM literature \citep{sae_lm, sae_protein}. For reference: GPT 4 \citep{gpt4} is estimated to have 1.8 trillion and Claude 3.5 Sonnet \citep{claude} 175B parameters. 
For the investigations here, we computed the AIS on the top 100 SAE features (not 200 as in the other experiments). The baseline results from \cite{sae_protein} take 200 features into consideration, those from \cite{sae_lm} look at 150, both w.r.t. the final layer embeddings.
To check the validity of the LLaVa captions, we run a randomized test on Gemma4-12b. By randomly shuffling the patch captions for every explained neuron, we arrive at a much lower (even negative) AIS.

\begin{wraptable}{r}{0.5\textwidth}
    \vspace{-13pt} 
    \centering
    \captionof{table}{\textbf{AIS performance across Gemma model scales, random shuffling control and external references.} Results use a fixed configuration (Vanilla SAE + DINO + CUB). Suffixes denote parameter counts in billions. For context, we provide external references from NLP* \citep{sae_lm} and PLM** \citep{sae_protein}, which use much larger models (GPT/Claude).}
    \label{tab:baselines}
    \resizebox{0.48\textwidth}{!}{
    \begin{tabular}{lc} 
        \toprule
        \textbf{Explainer/Simulator} & \textbf{Mean AIS}\\
        \midrule
        \multicolumn{2}{c}{\textit{Control (Random Caption Shuffling)}} \\
        \cmidrule(r){1-2} 
        Gemma4-12b & -0.0136 \\
        \midrule
        \multicolumn{2}{c}{\textit{Ours (Gemma Scaling Evaluation)}} \\
        \cmidrule(r){1-2} 
        Gemma4-2b & 0.2012 \\
        Gemma3-4b & 0.2327 \\
        Gemma4-4b & 0.3278 \\
        Gemma3-12b & 0.3313 \\
        \textbf{Gemma4-12b} &\textbf{0.3539} \\
        Gemma4-26b & 0.3269 \\
        \midrule
        \multicolumn{2}{c}{\textit{External Literature References (Larger Models)}} \\
        \cmidrule(r){1-2}
        *GPT 4\&3.5 (Text) & $\sim$ 0.15 \\
        **Claude 3.5 (Proteins) & $\sim$ 0.66 \\
        **Claude 3.5 (Amino-acids) & $\sim$ 0.63 \\
        \bottomrule
    \end{tabular}
    } 
    \vspace{-10pt} 
\end{wraptable}

\paragraph{Vision AIS Limitations. }

One of the contributions of this paper is the adaptation of the AIS routine from sequence-to-sequence modeling to vision. The central piece of transition is given by the captioning step (\autoref{algo:ais}, \autoref{fig:ais_pipeline}). While our experiments indicate that translating patch-level embeddings into text and delegating a lower-scale LLM to simulate a human explainer is a reasonable choice, achieves comparable results to baselines from the SAE literature and aligns with human explanations, 
several limitations are evident.

The cross-modality introduced into the AIS pipeline (vision embeddings $\rightarrow$ text captions) may be a form of lossy compression, as it is discarding visual information when captioning the patches. \autoref{fig:explanations_bar} in the Appendix gives an overview of the frequency of the top 20 words/concepts found in the Gemma-generated explanations for all configurations listed in \autoref{tab:sae_metrics}. Notice that most terms are rather generic, which, in turn, will impact the quality of the simulated activation scores. As mentioned above, the captioner's performance is highly relevant for the quality of the explanations. Likewise, the performance of the LLM Explainer and Simulator impacts the AIS. Given this chain of error propagation, it is, hence, not entirely clear to what extent a low AIS can be solely attributed to poor interpretability of SAE concepts.
    
The grid resolution on which the captioner is applied naturally impacts the level of detail from which the LLM explainer infers the explanation. While meant to equate the tokenization step in the AIS for sequence modeling, a static grid does not include any scene understanding whatsoever. As an alternative, other modern patching and captioning methods may be applied, that combine both steps into one and segment the image into irregular dynamic regions
- see \autoref{sec:extended_rel_work}. Furthermore, latest VLMs
would be able to solve the explanation and simulation tasks by directly analyzing the images, without needing captions at all. 
    


\section{Conclusion}\label{sec:conclusion}

In this work, we challenged the LRH on the basis of which SAEs are built. We have shown empirically that standard metrics designed to evaluate the interpretability of SAEs - whether from within or outside the concept dictionary - do not correlate with one another or with the AIS. We proposed an adaption of the AIS for vision models, previously shown in NLP to align with human explanations, and confirmed via a dedicated user study that our caption-based AIS procedure captures explanation quality as perceived by humans.

Our results support a growing line of work in explainability calling for more formalism in explanation methods~\citep{haufe_formal,haufe_paradigm}, deeper understanding of XAI metrics \citep{biessmann2021quality}, noise and uncertainty in explanations, supervision and ground-truth data for explainers and greater customization to task and audience \citep{luca_manifesto, teo_thesis}. For vision SAEs, new results show the benefits of steering the interpretability of SAE features via human-annotated concepts \citep{sae_concept_annot} or describing concepts beyond LRH-assumed linear separability \citep{projecting_assumptions}.


To conclude, more in-depth studies are needed on the efficiency and interpretability of SAEs, beyond standard LRH assumptions. We believe that, just like with any explanation method, there is no all-purpose SAE design guaranteed to deliver useful explanations across data and models. Task domain knowledge and reality-grounded metrics need to be interwoven into SAE training and evaluation.

\subsection*{AI use statement}


In this work, we used generative AI tools for providing feedback on research  methodology and experiments and implementing some of the methods.
We have not used generative AI tools for developing theoretical models or conceptual frameworks, formulating mathematical claims, providing critical ingredients for proving mathematical claims, proposing or refining hypotheses, assisting with translation,  supporting qualitative and thematic data analysis or interpreting results.
Generating synthetic data sets, writing of proofs, cleaning and reformatting datasets are not applicable to this work.
Additionally, we used generative AI tools for suggesting experimental parameters, debugging software code, analyzing existing literature, brainstorming and identifying gaps in the current literature. 
We have reviewed all AI-assisted work. LLM-generated code was verified and tested for correctness by 2 authors. The literature review and correctness of the notations and scientific claims were verified by all authors.
We take responsibility for the final content of this work,
including text, claims or artifacts produced with the aid of generative AI.





\subsection*{Reproducibility statement}

We have attached a link to our anonymized code repository, described our methodological approaches in great detail in the paper and provided references to and details about the data splits we used. 




\subsubsection*{Acknowledgments}
This research work was supported by the National Research Center for Applied Cybersecurity ATHENE. ATHENE is funded jointly by the German Federal Ministry of Research, Technology and Space and the Hessian Ministry of Science and Research, Arts and Culture.

\bibliography{iclr2027_conference}
\bibliographystyle{iclr2027_conference}

\newpage
\appendix
\section*{\textbf{Appendix}}

\section{Extended Related Work}\label{sec:extended_rel_work}

\paragraph{Sparse vs Distributed Coding.}

The debate between \textit{sparse and distributed coding} mechanisms has its roots in neuroscience research. On the one hand, the sparse coding hypothesis, often colloquially termed as the "Grandmother Cell" theory, posits that specific neurons exhibit extreme selectivity, firing exclusively in response to high-level, complex stimuli regardless of visual presentation or context. Empirical support for this localized representation stems from single-unit recordings in the human medial temporal lobe (MTL) \citep{QuianQuiroga2005, Kreiman2000}, which identified concept-specific neurons that fired selectively for specific individuals (such as Jennifer Aniston), places or objects. On the other hand, the distributed coding theory asserts that information is encoded across a population of neurons, not by individual cells. This framework accounts for the brain's capacity to represent a virtually unlimited array of stimuli using a limited number of neurons. This population-based view is widely supported across sensory and cognitive domains, as demonstrated by functional neuroimaging and electrophysiological studies \citep{Haxby2001, Norman2006}.

\paragraph{From Overcomplete Basis Functions to SAEs.}

The modern use of SAEs for MI in neural networks is rooted in foundational principles of overcomplete basis functions, sparse coding and Independent Component Analysis (ICA). The core premise that natural signals can be efficiently represented as sparse linear combinations of an overcomplete set of basis vectors was first established in the late 90's \citep{Olshausen1996, Olshausen1997}. Experiments showed that imposing a sparsity constraint on overcomplete representations yields localized, Gabor-like visual filters analogous to simple cells in the primary visual cortex. This computational paradigm shares strong theoretical links with ICA, as both seek to untangle complex distributions into independent, interpretable components. While ICA explicitly enforces statistical independence among the source components, sparse coding optimizes for structural sparsity where only a small subset of basis elements is active simultaneously. Later formalized into sparse dictionary learning frameworks using optimization algorithms like LASSO and iterative thresholding \citep{Jenatton2010}, these techniques provided a principled approach to learning dictionary matrices and sparse coefficients end-to-end. SAEs directly adapt this dictionary learning paradigm to modern Deep Learning architectures. By introducing an explicit sparsity penalty to the latent bottleneck, e.g. through $L_1$ regularization, SAEs force the network to reconstruct dense internal activations through a small combination of monosemantic latent directions.

\paragraph{Localized Image Captioning. } 

In order to strike a bridge between sequence modeling and vision w.r.t. automatic interpretability evaluation, our work introduces a captioning step. Conventional VLMs, such as BLIP \citep{blip} and ViT-GPT2 \citep{vitgpt2}, are predominantly trained on image-level descriptions such as COCO \citep{coco}, resulting in a significant object-centric inductive bias that favors holistic descriptions over fine-grained visual textures. While CLIP-based methods \citep{clip,selfsupervisedimagecaptioningclip} attempt to align visual and textual embeddings, they typically operate at a global scale. Recent instruction-tuned VLMs like InstructBLIP \citep{instructblip} and LLaVA \citep{llava} introduce spatial/context awareness via prompting.

Other recent work has shifted towards detailed localized captioning (DLC) and dense captioning. Frameworks such as Segment and Caption Anything (SCA) \citep{sca} and the Describe Anything Model (DAM) \citep{dam} leverage the Segment Anything Model (SAM) \citep{sam2,sam3} to generate context-aware descriptions for arbitrary regions and masks. \cite{patchioner} propose the Patch-ioner framework, which adopts a patch-centric paradigm to aggregate patch representations into coherent descriptions of non-contiguous areas. 

\paragraph{Zero-Shot Automated Vision Neuron Description. }

Within the rising need for model audition tools, research on MI has developed zero-shot pipelines for assigning natural language descriptions directly to individual neurons in vision networks. As an early attempt, Network Dissection \citep{netdissect2017} measures how closely a neuron's activations align with concepts from a labeled, predetermined concept set, which ties its descriptions to the coverage and granularity of that fixed annotation vocabulary. MILAN \citep{milan} removes the need for a pre-annotated label set by searching for an open-ended natural language string that maximizes pointwise mutual information with the image regions where a neuron is active. Its descriptions are validated through agreement with human-written descriptions. CLIP-Dissect \citep{clip_dissect} instead uses a VLM to score neuron activations against a user-specified concept set without requiring labeled probing data.


\section{SAE Training}\label{sec:sae_training}

\paragraph{Training Objective. }

SAEs are trained jointly to minimize the error between the original patch embeddings $e$ and the reconstructed ones $\hat{e}$, while maintaining a sparse latent representation $z$. For readability, we omit the indexes $i, j$ here. The general training loss is formulated as:
\begin{equation}\label{eq:sae_loss}
    \mathcal{L} = \|e - \hat{e}\|_2^2 + \lambda \mathcal{R}(z) + \alpha \mathcal{L}_{aux}
\end{equation}
where:
\begin{itemize}[nolistsep,leftmargin=.3in]
    \item $\|e - \hat{e}\|_2^2$ is the reconstruction Mean Squared Error (MSE),
    \item $\mathcal{R}(z)$ is a penalty function promoting sparsity in the latent activations (e.g., via $L_1$ or $L_0$ mechanisms), with $\lambda$ as a hyperparameter steering the trade-off between reconstruction fidelity and sparsity (usually set between $10^{-2}$ and $10^{-3}$ \citep{bricken2023monosemantic}),
    \item $\mathcal{L}_{aux}$ is an auxiliary loss term utilized to recycle inactive or "dead" codes, scaled by a coefficient $\alpha$ (typically chosen as $\frac{1}{32}$ \citep{topk_sae}).
\end{itemize}

\paragraph{SAE Architectures. }

Recent literature has introduced several variants of the SAE architecture, that primarily differ in the $\sigma$ projection and loss structures. We briefly describe here the ones relevant for our experiments:

\begin{itemize}[nolistsep,leftmargin=.3in]
    \item \textbf{ReLU (Vanilla) SAEs} \citep{bricken2023monosemantic}: The baseline SAE utilizes a standard ReLU to project the affine transformation of $e$ in \autoref{eq:sae_enc}. To enforce sparsity, it uses an $L_1$ penalty, setting $\mathcal{R}(z) = \|z\|_1$ and $\alpha = 0$.

    \item \textbf{TopK SAEs} \citep{topk_sae, batchtopk_sae}: A standard TopK SAE retains only the $K$ largest activated codes per sample, explicitly zeroing out the rest: $\sigma(\cdot) = \text{TopK}(\cdot)$. This replaces the explicit sparsity penalty ($\lambda = 0$), allowing direct control over the $L_0$-norm. Because larger SAEs are prone to "dead features" (neurons that completely stop activating over multiple iterations), TopK SAEs often incorporate an auxiliary loss $\mathcal{L}_{aux}$ that models reconstruction error using the top $K_{aux}$ dead features to encourage them back into the  support of $z$ (the set of active neurons).

    \item \textbf{MP-SAEs} \citep{matching_pursuit}: The authors of MP-SAE discuss the limits of LRH, given by concepts that are non-linearly accessible or hierarchical. In this respect, they propose unrolling the Matching Pursuit algorithm \citep{mp} into a multi-step encoder. MP-SAE shares the same matrix as encoder and decoder
    : $W_{enc} = W_{dec}^T = D$. Starting from an initial "residual" $e - b_{pre}$, the model iteratively selects the dictionary atom that maximizes the inner-product with the residual. This inner-product is the single sparse feature in $z$ used to update the reconstruction approximation and refine the residual in the current step $t \leq T$. In our nomenclature, this equates to applying a Top-1 $\sigma$-projection $T$ times, resulting in a representation $z$ of sparsity $\|z\|_0^{} \leq T$. As for the loss, we train MP-SAE with a simple MSE loss ($\lambda = \alpha = 0$)\footnote{In the original paper \citep{matching_pursuit}, the authors add a reanimation term to the MSE to reactivate features in danger of dying out.}.
\end{itemize}

\paragraph{Hyperparameters. }

Following the recommendations from \cite{universal_sae} and \cite{matching_pursuit}, we train the Vanilla and TopK SAE for 30 epochs and the MP-SAE for 10 epochs with a cosine schedule, warming up the learning rate from $10^{-6}$ to $5 \cdot 10^{-4}$ for the first 5 epochs, then reducing it back to $10^{-6}$ by the final epoch. For this, we coupled a \textit{LinearLR} with a \textit{CosineAnnealingLR} in a \textit{SequentialLR} from PyTorch. The optimizer has a fixed weight decay of $10^{-5}$. Regarding the layers in the SAEs, the encoder is a linear transformation given by $W_{enc}$ in \autoref{eq:sae_enc}, along with Batch Normalization (BN). While this introduces batch-dependent statistics during training, at inference the encoder uses the fixed running mean and variance accumulated over training, so each patch embedding maps deterministically to a fixed latent code at test time. We include BN because, in preliminary trials, omitting it from the encoder consistently degraded MS, OOD Score, Coherence and Connectivity across configurations, suggesting it plays a stabilizing role in training useful dictionaries in our setting.

The dictionary consists solely of the matrix $W_{dec}^T$ and has size $m = 16 \cdot 768 = 12288$. We arrive at this number by multiplying the largest embedding dimension $p$ over the vision backbones considered (ViT and SigLIP have 768, DINO has 384) by 16. The expansion factor for DINO is, therefore, 32. For the TopK-SAEs, we chose $K$ in the top-K-projection such that 20\% of the latent features remain active. We set the number of iterations $T$ in MP-SAE to 100, which leads to an average sparsity of $\frac{100}{12288} \approx 0.0081$ (\autoref{tab:sae_metrics}).

We apply the SAEs on top of the final embedding layer of the vision backbones, as well as on various intermediate layers (\autoref{tab:sae_metrics_intermed}). Hence, $e_j$ and $\tilde{e}_j$ will refer to the final or any other intermediate embeddings, depending on the context. As we are not interested here in measuring downstream task performance, we did not further train a task-solver head on top of the SAE features.

For the auxiliary loss term in \autoref{eq:sae_loss} we set $\alpha = \frac{1}{32}$, as recommended in \cite{topk_sae}, and $K_{aux} = \text{min}(512, \# \text{dead codes})$ and insert this term only for training TopK-SAEs. As for Vanilla SAEs, we are setting the $L_1$ coefficient $\lambda = 10^{-5}$.

For carrying out the experiments, we implemented SAE architectures from the \textit{overcomplete} library \citep{overcomplete} and tracked all the different configurations via \textit{hydra} \citep{hydra}. For more details, we refer the reader to our repository. Note that some authors, e.g. \citep{batchtopk_sae}, leave out the explicit term $b_{pre}$ from their encoder's signature, since $-W_{enc}b_{pre}$ can be merged into a common bias term $b =: b_{enc}$. The encoder implementation in the \textit{overcomplete} library \citep{overcomplete} also leaves it out.

\section{Other SAE Evaluation Metrics}\label{sec:other_metrics}

\paragraph{$R^2$ Score. } This is the standard coefficient of determination, which we report as a measure of the reconstruction quality: 1 means perfect reconstruction $e = \hat{e}$, 0 means that the reconstruction is as good as the mean.

\paragraph{$L_0$ Sparsity. } We compute the ratio of non-zero SAE activations as:
\begin{equation}\label{eq:sparsity}
    \text{Sparsity} = \frac{1}{nsm} \|Z\|_0^{}
\end{equation}

\paragraph{Monosemanticity Score. } 

In order to evaluate the purity of SAE neurons globally, we implemented the Monosemanticity Score (MS) \citep{sae_monosemantic}. For each neuron, this score quantifies how similar the images are, that highly activate that neuron. To compute the MS, we first construct a similarity matrix 
\begin{equation}
    S := \tilde{E} \cdot \tilde{E}^T 
\end{equation}
based on all the pairwise cosine similarities between $L_2$-normalized image embeddings\footnote{\cite{sae_monosemantic} use a different vision embedder for computing the matrix $S$ than the embedder explained by their SAEs. We follow this idea and compute $S$ with ViT when DINO is explained, with SigLIP when ViT is explained and with DINO when SigLIP is explained.} $\tilde{e}_i \in \mathbb{R}^p, \forall 1 \leq i \leq n$ of the considered dataset. The min-max normalized image activations $\tilde{a}^{(k)}$ of the $k$-th neuron are also cross-multiplied, to yield a relevance matrix $R^{(k)}$ of the shared neuron activation of each image pair: 
\begin{equation}
    R^{(k)} := \tilde{a}^{(k)} (\tilde{a}^{(k)})^T. 
\end{equation}
The MS of the $k$-th neuron is, therefore, the average pairwise similarity weighted by the relevance scores (excluding the self-similar pairs):
\begin{equation}
    MS^{(k)} := \frac{1}{n(n-1)} \sum_i^n \sum_{l \neq i}^n \tilde{a}_i^{(k)} \tilde{a}_l^{(k)} (S)_{i,l},
\end{equation}
To evaluate this metric efficiently at scale without computing an explicit $n \times n$ relevance matrix for each neuron, we reformulate the pairwise summation as a quadratic form. Because the original formulation explicitly excludes self-similarity pairs ($i = l$), we subtract the diagonal elements, where the self-similarity $S_{i,i} = 1$ and the joint relevance reduces to $(\tilde{a}_i^{(k)})^2$:
\begin{equation}
    MS^{(k)} := \frac{1}{n(n-1)} \left( (\tilde{a}^{(k)})^T S \tilde{a}^{(k)} - \sum_{i=1}^n (\tilde{a}_i^{(k)})^2 \right).
\end{equation}


\paragraph{OOD Score. } To assess how grounded the learned dictionary is in real data, we compute the Out-of-Distribution (OOD) Score \citep{archetypal_sae}, which measures the deviation of each dictionary atom from the closest patch-level embedding it could represent:
\begin{equation}
    \text{OOD Score} = 1 - \frac{1}{m} \sum_{k=1}^{m} \max_j \langle d_k, e_j \rangle,
\end{equation}
where both $d_k, e_j \in \mathbb{R}^p$ are $L_2$-normalized. A score of 0 indicates that every atom exactly matches an existing activation (pure lookup behavior), while higher values indicate that atoms drift away from the real data manifold. We compute this metric using patch-level embeddings, since this is the feature space the SAE is actually trained to reconstruct; evaluating against image-level embeddings would test dictionary atoms against a distribution the encoder never observes.

\paragraph{Coherence. } We measure redundancy within the dictionary via Coherence \citep{archetypal_sae}, the maximum pairwise cosine similarity between distinct dictionary rows:
\begin{equation}
    \text{Coherence} = \max_{k \neq l} \left| d_k^\top d_l \right|,
\end{equation}
with each $d_k$ constrained to the unit $L_2$ sphere. Low coherence indicates that dictionary atoms span diverse, near-independent directions, which is desirable for disentangled representations; high coherence indicates that some rows encode near-duplicate features, reducing the effective capacity of the learned basis.

\paragraph{Connectivity. } To characterize how concepts combine in the learned codes, we measure Connectivity \citep{archetypal_sae} as the fraction of distinct image-level concept pairs that co-activate at least once across the dataset:
\begin{equation}
    \text{Connectivity} = \frac{1}{m^2} \| \tilde{Z}^\top \tilde{Z} \|_0^{}.
\end{equation}
A high connectivity score indicates that a broad range of concepts can be meaningfully combined to reconstruct embeddings, while low connectivity indicates a more modular representation in which only a small subset of concepts co-occurs. 
We note that, as opposed to the original implementation in \citep{archetypal_sae}, we remove the subtraction from 1, deeming the metric more intuitive this way.

\section{Max Filter}\label{sec:max_filter}

We decide which SAE features to explain based on the following filter (steps 3 and 4 are inspired by \cite{sae_protein}):
\begin{enumerate}[nolistsep,leftmargin=.3in]
    \item In order to match the captioner's $4 \times 4$ grid, each $a^{(k)} \in \mathbb{R}^{ns}$ is average-pooled into $\textbf{a}^{(k)} \in \mathbb{R}^{n \cdot 16}$. Note that $s$ is not always a multiple of 16 for all considered embedders, but by applying an adaptive average 2D-pooling, we ensure a $4 \times 4$ pooled grid for consistency.

    \item Compute the mean pooled activations per image per neuron for later use in ranking (Step 4):
    $\overline{\textbf{a}}_i^{(k)} = \frac{1}{16} \sum_{j=16(i-1)+1}^{16i} \textbf{a}_j^{(k)} \in \mathbb{R}$. 
    
    \item Filter out insufficiently activated SAE features by minimum firing frequency, computed at the level of individual pooled patch activations $\textbf{a}_j^{(k)}$ (prior to any per-image averaging):
    $$f^{(k)} = \frac{1}{16n} \sum_{j=1}^{16n} \mathbbm{1}\left[\textbf{a}_j^{(k)} > 0\right].$$
    $\mathbbm{1}[\cdot]$ denotes the indicator function. We only retain features with $f^{(k)} > \frac{2 \cdot 10}{n}$, i.e. requiring more than $2 \cdot 10 \cdot 16 = 320$ individual patch-level activations to be positive - twice the number of activating patches we would need if 10 images fired across the full $4\times4$ grid.
    
    \item Rank remaining features by the averaged mean pooled activations across all images ($\overline{\textbf{a}}^{(k)} = \frac{1}{n} \sum_i \overline{\textbf{a}}_i^{(k)}$) and keep the top 200 for interpreting.
\end{enumerate}

\clearpage
\section{Extra AIS Values}\label{sec:extra_ais}

\begin{table}[htbp]
\centering
\caption{\textbf{Extra AIS values.} The \textbf{p}-value stems from testing the null hypothesis that the predicted and ground truth distributions of the neuron activations are uncorrelated; here we report the percent of the neurons that have this \textbf{p}-value less than 0.05. \textbf{Errors} occur whenever the simulator predicts a different number of SAE activations than in the ground truth. \textbf{NaNs} occur whenever $(\hat{A}_{\text{pred}}, A_{\text{truth}})$ is a zero-pair. Both cases are not considered in the AIS computation.}
\label{tab:ais_extra_metrics}
\resizebox{0.8\textwidth}{!}{
\begin{tabular}{lllccccc}
\hline
\textbf{SAE} & \textbf{Embedder} & \textbf{Dataset} & \textbf{Mean AIS} & \textbf{Median AIS} & \textbf{\% p < 0.05} & \textbf{Errors} & \textbf{NaNs} \\
\hline
Vanilla & DINO   & CUB200      & 0.3493 & 0.3477 & 80.5 & 0 & 1 \\
        &        & ImageNet100 & 0.2801 & 0.2962 & 73.5 & 0 & 1 \\
        &        & Caltech     & 0.2049 & 0.1960 & 66.0 & 0 & 0 \\
\cline{2-8}                            
        & ViT    & CUB200      & 0.1621 & 0.1699 & 61.0 & 0 & 0 \\
        &        & ImageNet100 & 0.2105 & 0.1887 & 66.0 & 0 & 2 \\
        &        & Caltech     & 0.1510 & 0.1748 & 59.5 & 0 & 0 \\
\cline{2-8}                            
        & SigLIP & CUB200      & 0.3518 & 0.3668 & 78.5 & 0 & 0 \\
        &        & ImageNet100 & 0.3545 & 0.3774 & 81.0 & 0 & 0 \\
        &        & Caltech     & 0.2522 & 0.2670 & 73.5 & 0 & 0 \\
\hline\hline                           
Top-K   & DINO   & CUB200      & \tikzmark{topk_start}0.1619 & 0.1439 & 52.5\tikzmark{topk_end} & 0 & 1 \\
        &        & ImageNet100 & 0.0820 & 0.0803 & 38.0 & 0 & 6 \\
        &        & Caltech     & 0.0636 & 0.0563 & 36.0 & 0 & 1 \\
\cline{2-8}                            
        & ViT    & CUB200      & 0.0226 & 0.0035 & 43.0 & 1 & 3 \\
        &        & ImageNet100 & 0.0505 & 0.0529 & 56.7 & 1 & 3 \\
        &        & Caltech     & 0.0477 & 0.0433 & 45.0 & 0 & 3 \\
\cline{2-8}                            
        & SigLIP & CUB200      & 0.1458 & 0.1310 & 50.2 & 1 & 2 \\
        &        & ImageNet100 & 0.0587 & 0.0559 & 41.0 & 0 & 8 \\
        &        & Caltech     & 0.0779 & 0.0909 & 44.5\tikzmark{topk_bottom_right} & 0 & 1 \\
\hline\hline                           
MP      & DINO   & CUB200      & 0.3476 & 0.3302 & 79.0 & 0 & 0 \\
        &        & ImageNet100 & 0.4332 & 0.4619 & 80.0 & 0 & 0 \\
        &        & Caltech     & 0.3763 & 0.3923 & 77.0 & 0 & 0 \\
\cline{2-8}                            
        & ViT    & CUB200      & 0.2817 & 0.2858 & 72.5 & 0 & 7 \\
        &        & ImageNet100 & 0.4222 & 0.4879 & 78.0 & 0 & 8 \\
        &        & Caltech     & 0.3834 & 0.3913 & 78.5 & 1 & 7 \\
\cline{2-8}                            
        & SigLIP & CUB200      & 0.2750 & 0.2701 & 55.0 & 0 & 36 \\
        &        & ImageNet100 & 0.3213 & 0.2971 & 63.5 & 0 & 12 \\
        &        & Caltech     & 0.2764 & 0.2838 & 57.5 & 0 & 33 \\
\bottomrule
\end{tabular}%
\begin{tikzpicture}[overlay, remember picture]
    \draw[red, very thick] 
        ([shift={(-3pt, 8pt)}]pic cs:topk_start) 
        rectangle 
        ([shift={(3pt, -3pt)}]pic cs:topk_bottom_right);
\end{tikzpicture}
} 
\end{table}

We note that the proportion of neurons reaching statistical significance (\% 
p < 0.05) is closely tied to the underlying strength of the predicted-vs-ground-truth correlation: configurations with markedly lower Mean and Median AIS - most notably Top-K SAEs, which score substantially below Vanilla and MP SAEs across all embedders and datasets - also show the lowest significance rates. Given that significance is computed per neuron over a comparatively small number of held-out images (namely, 10), weaker true correlations are disproportionately likely to fall below the significance threshold even when the simulator captures a genuine, if modest, relationship. We, therefore, interpret the lower significance rates for these configurations as consistent with the comparatively weak AIS scores already observed for Top-K SAEs.

\section{Examples LLM Explanations and Simulations}\label{sec:ex_llm}

\begin{figure}
    \centering
    \includegraphics[width=0.9\linewidth]{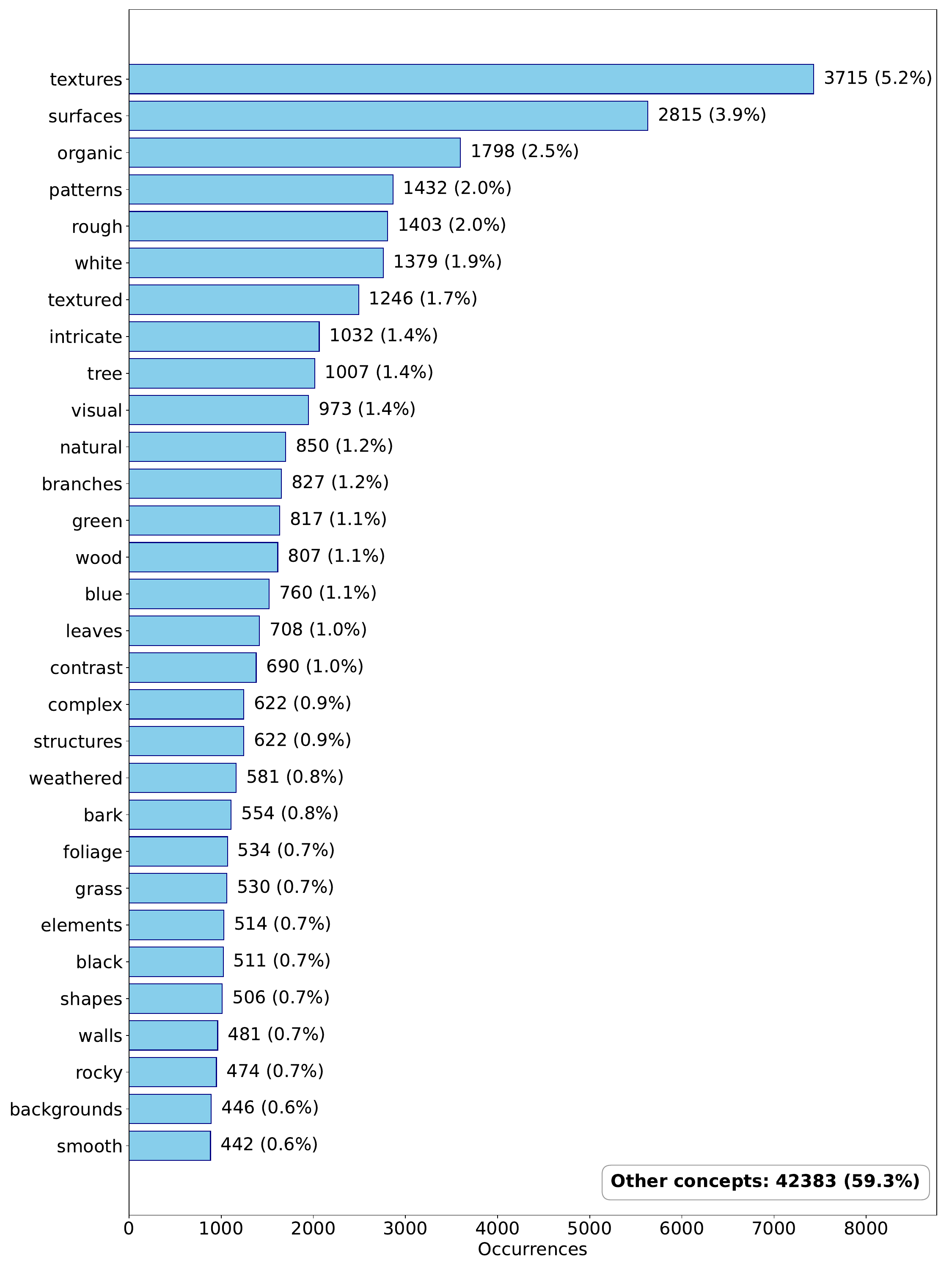}
    \caption{Frequency of the top 20 concepts/words in all the explanations generated in our AIS experiments (\autoref{tab:sae_metrics}).}
    \label{fig:explanations_bar}
\end{figure}

\autoref{fig:output_ais_explain} and \autoref{fig:output_ais_simulate} show truncated examples of the prompts and output files generated during the AIS pipeline for a Vanilla SAE on DINO embeddings and CUB data, where Gemma4-12b was both explainer and simulator. \autoref{fig:explanations_bar} gives an overview of the frequency of common concepts in all the LLM-generated explanations. 

\begin{figure}[t]
\centering
\begin{tcolorbox}[colback=gray!5!white,colframe=gray!75!black,fontupper=\small\ttfamily,boxrule=0.5pt,arc=2pt,left=6pt,right=6pt,top=6pt,bottom=6pt]
\textbf{+++ NEURON 4762 (8/200) +++}\\
\\
\textbf{PROMPT FOR EXPLANATION:}\\
You are an expert in interpretable AI analyzing a neuron in a vision model.\\
I will provide a list of image patch descriptions along with the 'Activation' score of the neuron for each patch.\\
Activation values range from 0 to 10.\\
High activation (above 5.0) means the neuron fired strongly. Low activation (near 0.0) means it did not care.\\
\\
\textbf{DATA:}\\
\textit{PATCH 0:  (ACTIVATION: 5.00)} - CAPTION: The image features a close-up of a green, grassy field. The grass appears to be lush and vibrant, with a mix of short and long blades. The field is filled with a variety of shapes and textures, creating a visually appealing and natural landscape.\\
(...)\\
\textit{PATCH 4:  (ACTIVATION: 6.00)} - CAPTION: The image features a close-up of a blue and white sky. The blue color dominates the sky, while the white clouds add texture and depth to the scene. The sky appears to be cloudy, with the clouds taking on various shapes and sizes. The overall atmosphere of the image is serene and peaceful.\\
\textit{PATCH 5:  (ACTIVATION: 3.00)} - CAPTION: The image features a close-up of a bird's wing, showcasing its feathers. The feathers exhibit a mix of textures, shapes, and colors, including a combination of gray, brown, and white hues. The wing's structure is intricate, with a pattern of overlapping feathers that create a unique visual effect. The close-up view highlights the bird's wing as a fascinating and intricate part of its anatomy.\\
(...)\\
\textit{PATCH 26: (ACTIVATION: 3.00)} - CAPTION: The image features a close-up of a bird's feathers, showcasing the intricate patterns and textures of its plumage. The feathers exhibit a mix of colors, including brown, black, and white, creating a visually striking scene. The bird's feathers are arranged in various shapes, such as curved lines and irregular patterns, which contribute to the overall beauty of the bird's appearance.\\
(...)\\
\textit{PATCH 79: (ACTIVATION: 6.00)} - CAPTION: The image features a close-up of a leaf with a green color. The leaf has a textured surface, with visible veins and a slightly rough texture. The leaf is curled, giving it a unique shape. The close-up view highlights the intricate details of the leaf, making it an interesting and visually appealing subject.\\
\\
\textbf{TASK:}\\
Identify the common visual concept (e.g., shape, texture, color, object part) that causes high activation.\\
Provide a concise final answer in one sentence.\\
\\
\textbf{EXPLANATION:}\\
The neuron primarily responds to green vegetation, specifically focusing on the textures, veins, and shapes of leaves, grass, and plants.
\end{tcolorbox}
\caption{Example of an explanation generated by Gemma4-12b.}
\label{fig:output_ais_explain}
\end{figure}

\begin{figure}[t]
\centering
\begin{tcolorbox}[colback=gray!5!white,colframe=gray!75!black,fontupper=\small\ttfamily,boxrule=0.5pt,arc=2pt,left=6pt,right=6pt,top=6pt,bottom=6pt]
\textbf{+++ NEURON 4762 (8/200) +++}\\
\\
\textbf{PROMPT FOR SIMULATION:}\\
You are an expert in interpretable AI. You have been given a specific visual concept that represents what a neuron (or Sparse Autoencoder feature) detects in a vision model. Your task is to act as a simulator: given a description of a new image patch, predict how strongly this neuron will activate based on the provided concept.

Activation Scale (whole numbers):\\
- 10.0: Very strong activation (patch matches the concept perfectly).\\
- 5.0 to 10.0: Strong activation (patch matches the concept well).\\
- 0.0 to 5.0: Moderate activation (patch partially matches or contains the concept)\\
- 0.0: No activation (concept entirely absent).\\
\\
\textbf{DATA:}\\
\textit{PATCH 0:}    CAPTION: The image features a close-up of tall, green grass with a mix of light and dark shades. The grass appears to be slightly blurred, giving it a soft and natural texture. (...)\\
(...)\\
\textit{PATCH 2:}    CAPTION: The close-up image features a bird with a brown and white color pattern. The bird's head is prominently visible, and its beak is open. (...)\\
(...)\\
\textit{PATCH 37:}   CAPTION: The image features a close-up of a yellow and black bird perched on a branch. The bird's feathers exhibit a mix of yellow and black colors, creating a visually striking pattern. (...)\\
(...)\\
\textit{PATCH 159:} CAPTION: The image features a close-up of a body of water, possibly a lake or a river. The water appears to be calm and reflective, with a blue hue. (...)\\
\\
\textbf{EXPLANATION:}\\
The neuron primarily responds to green vegetation, specifically focusing on the textures, veins, and shapes of leaves, grass, and plants.\\
\\
\textbf{TASK:}\\
Based on the explanation and the captions provided, predict the activation score (integer) for each patch.

STRICT OUTPUT RULES:\\
1. Output format: PATCH [ID]: [ACTIVATION]\\
2. Do NOT repeat the caption.\\
3. Do NOT provide reasoning or extra text.\\
\\
\textbf{AIS Results:}\\
AIS (Pearson r): \textbf{0.7609}\\
P-value: 1.7208e-31\\
Comparison (GT vs Predicted):\\
  Patch 0:   Ground Truth   5.0 | Simulator   9.0\\
  Patch 1:   Ground Truth   5.0 | Simulator   8.0\\
  Patch 2:   Ground Truth   4.0 | Simulator   4.0\\
  Patch 3:   Ground Truth   5.0 | Simulator   9.0\\
  Patch 4:   Ground Truth   5.0 | Simulator   9.0\\
  Patch 5:   Ground Truth   5.0 | Simulator   5.0\\
  (...)\\
  Patch 155: Ground Truth   1.0 | Simulator   0.0\\
  Patch 156: Ground Truth   2.0 | Simulator   0.0\\
  Patch 157: Ground Truth   1.0 | Simulator   0.0\\
  Patch 158: Ground Truth   2.0 | Simulator   0.0\\
  Patch 159: Ground Truth   2.0 | Simulator   0.0
\end{tcolorbox}
\caption{Example of a neuron simulation done by Gemma4-12b.}
\label{fig:output_ais_simulate}
\end{figure}

\clearpage
\clearpage
\clearpage
\section{User Study}\label{sec:user_study}

\begin{figure}[htbp]
    \centering
    \begin{subfigure}[b]{0.48\textwidth}
        \centering
        \includegraphics[width=\linewidth]{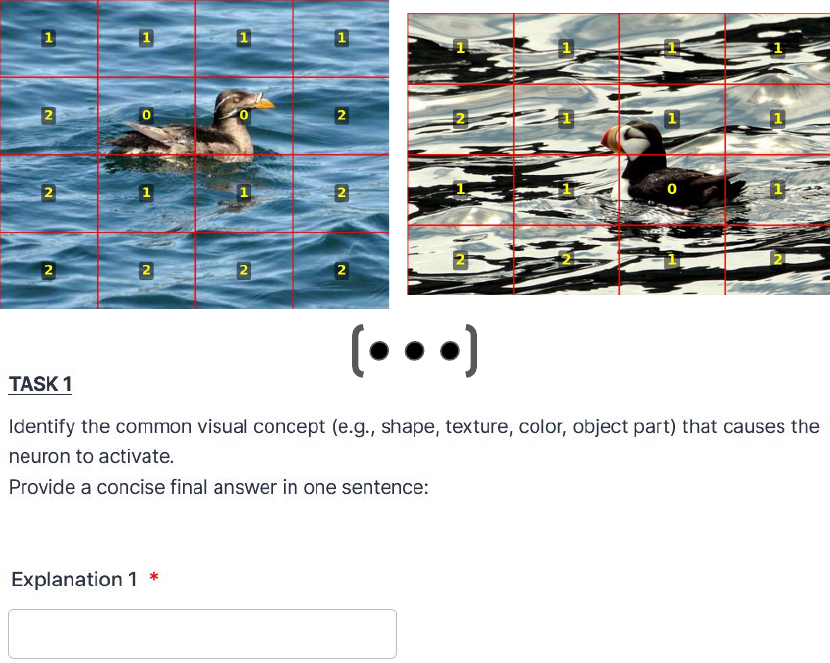}
        \caption{Example explanation task (for the "water" neuron)}
    \end{subfigure}
    \hfill
    \begin{subfigure}[b]{0.48\textwidth}
        \centering
        \includegraphics[width=\linewidth]{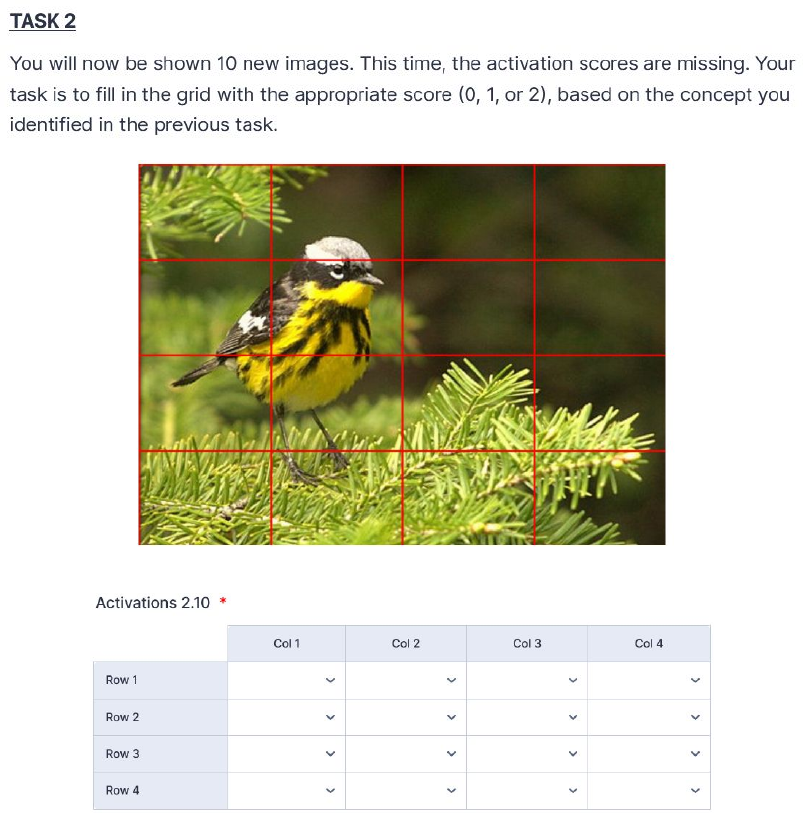}
        \caption{Example simulation task}
    \end{subfigure}
    \caption{Excerpts from the user study interface.}
    \label{fig:user_ui}
\end{figure}

The experiment was approved by our ethics board and included 21 initial participants. Our experiment was anonymous and hosted on the \textit{Jotform} platform\footnote{\href{https://www.jotform.com}{https://www.jotform.com}}. The subjects were informed at the very beginning that their annotations would solely be used for research purposes. Excerpts from the user interface can be found in \autoref{fig:user_ui}.

We selected two of the topmost activating MP SAE neurons (trained on DINO embeddings for CUB data): one focusing on (leafless) tree branches, one on bodies of water. Just like in our automated explanation pipeline, the users were first asked to find the concept the neurons were firing for, based on 5 examples of top activating images along with the patch activation scores. The subjects were shown these scores directly on top of the images, not together with patch captions. Also, in order to reduce cognitive load, we mapped the original integer scores from $[0, 10]$ to $\{0, 1, 2\} = $ \{low, medium, high\} and restricted the experiment to only two neurons, as this alone took our participants 25 minutes on average to complete. In the simulation task, they were asked to annotate 10 top-random (5 top activating + 5 random) examples with these scores (16 scores per image). The images in both tasks and neurons were randomly shuffled for every user.

After collecting the survey responses, we removed the samples where the user's explanations for both neurons were very unspecific or missing, e.g. \textit{"I don't know"} or \textit{"an object part"}. This led to a total of 16 valid users, whose annotations were considered for the statistics in \autoref{sec:results}. The users' responses can be found in our repository. Additionally, 3 (out of 224) user non-zero annotations in the simulation task with zero variance, i.e. all-ones or all-twos, were also removed from the average AIS values per image in \autoref{fig:ais_llm_users}, as the SAE ground-truth also did not contain any of this kind.

\begin{figure}[t]
\centering
\begin{tcolorbox}[colback=gray!5!white,colframe=gray!75!black,fontupper=\small\ttfamily,boxrule=0.5pt,arc=2pt,left=6pt,right=6pt,top=6pt,bottom=6pt]
\textbf{Explanation 1 (Neuron "branches"):}\\
\\
\textbf{Gemma:}\\
The neuron consistently fires strongly on images featuring the rough texture, knots, twists, and organic shapes of tree branches, particularly when they are shown in close-up.\\
\\
\textbf{Users:}\\
- The activation is caused by detection of tree branches.\\
- The neuron is triggered by tree branches, particularly those with a specific branch structure.\\
- The common visual concepts are the texture and the colour\\
- Structure: branched branches; Colors: blue background and brown branches\\
- tree branches\\
- The activation seems to be linked to the branches and the area surrounding the bird's body – and could therefore relate to either the animal's habitat or its size.\\
- background/tree\\
- Twigs/branches of the tree, shape: cylindrical, color: some form of brown\\
- The most common visual concept is the branches the bird sits in.\\
- The texture and shape of the tree components, i.e. its bark and branches\\
(...)\\
\\
\textbf{Explanation 2 (Neuron "water"):}\\
\\
\textbf{Gemma:}\\
The neuron strongly activates on the visual concept of water surfaces, specifically focusing on the textures of ripples, waves, and reflections on bodies of water like lakes or oceans.\\
\\
\textbf{Users:}\\
- Activation is triggered by the detection of a water surface.\\
- The neuron is triggered by the texture and wave patterns of the water surface, with stronger activation for more pronounced wave structures.\\
- Between 0 and 1 there is a clear difference in colour and texture, maybe also shape. However I cannot tell the difference between 1 and 2\\
- the structure of water, with its movement and the way it reflects light\\
- water, soft shapes\\
- Again the model seems to pay attention at the animal‘s surroundings, maybe also at certain textures/patterns in that surrounding.\\
background/water\\
- color: blue, grey, green, texture: rippled, object part: water surface\\
- The model pays most attention to the water, especially with tiny waves.\\
- It seems as if the neurons focus on the water color\\
(...)
\end{tcolorbox}
\caption{Gemma and example user explanations for the two surveyed neurons.}
\label{fig:user_explanations}
\end{figure}

\clearpage
\section{Ablation Study: Layer-wise MS and AIS}\label{sec:ablation}


\begin{wraptable}{r}{0.5\textwidth}
    \vspace{-13pt} 
    \centering
    \caption{\textbf{Earlier Layers Are More Interpretable and Raw Embeddings Have a Higher MS Score than SAE Features, but a Lower AIS.} The results here are computed on DINO embeddings for the CUB dataset. The left table half is computed with a Vanilla SAE. The AIS explanations and simulations are generated with Gemma4-12b. The NaNs in the right half stem from zero-pairs $(\hat{A}_{\text{pred}}, A_{\text{truth}})$; these were not included in the AIS computation.}
    \label{tab:sae_metrics_intermed}
    \resizebox{0.5\textwidth}{!}{
    \begin{tabular}{lcc||cc}
    \toprule
    & \multicolumn{2}{c||}{\textbf{SAE}} & \multicolumn{2}{c}{\textbf{No SAE}} \\
    \cline{2-3} \cline{4-5}
    \textbf{Layer} & \textbf{MS} & \textbf{AIS} & \textbf{MS} & \textbf{AIS} \\
    \hline
    final & 0.0500 & 0.3493 & 0.0602 & 0.2764 \quad (2 NaNs) \\
    8     & 0.0447 & 0.2968 & \textbf{0.0612} & 0.2376 (127 NaNs) \\
    5     & 0.0552 & \textbf{0.3965} & 0.0609 & 0.2768 (167 NaNs) \\
    2     & \textbf{0.0605} & 0.2994 & 0.0610 & \textbf{0.2971} (181 NaNs) \\
    \bottomrule
    \end{tabular}%
    } 
\end{wraptable}

We looked at the MS and AIS of DINO embeddings on the CUB dataset also on various intermediate layers and compared the interpretability of the Vanilla SAE features w.r.t. to these two metrics to the interpretability of the raw DINO embeddings (\autoref{tab:sae_metrics_intermed}). We apply the same filter described in \autoref{sec:max_filter} on the raw embeddings. The explanations and simulations are generated with Gemma4-12b. Firstly, we note that the highest MS and AIS values are not in the final embeddings, but on previous layers. This is consistent with other empirical results from the SAE literature, e.g. \citep{sae_lm, sae_monosemantic}, which suggests that earlier DNN layers encode more monosemantic concepts. 
Recent work on the phenomenon of \textit{neural collapse} \citep{neural_collapse} supports this finding, while newer embedder architectures aim at dealing with this problem \citep{dino3}.

Secondly, we observe that input embeddings across all layers have a higher MS score than their corresponding SAE features, yet lower AIS values. This inconsistency between MS and AIS seconds the lack of correlation documented in \autoref{fig:corr_mat}. 

Thirdly, notice the increasing frequency of NaNs (\textit{Not a Number}) towards the earlier layers for the No-SAE case. We verified that all NaNs arise from degenerate pairs $(\hat{A}_{\text{pred}}, A_{\text{truth}}) = (0, 0)$. On the one hand, this is evidence that the LLM simulator is consistent at the floor of the activation range and does not hallucinate non-zero activations. On the other hand, this on its own would not be sufficient evidence of a well-calibrated simulator, since a naive simulator that always predicts 0 would also pass this test; however, we confirmed that predictions on the non-degenerate pairs show meaningful variance rather than defaulting to zero.

\end{document}